\documentclass[fleqn,10pt]{wlscirep}
\usepackage[utf8]{inputenc}
\usepackage[T1]{fontenc}
\usepackage{listings}
\usepackage{tabularx}
\usepackage{booktabs}
\usepackage{xcolor}

\title{Deep Fast Machine Learning Utils: A Python Library for Streamlined Machine Learning Prototyping}

\author[1,2,*]{Fabi Prezja}
\affil[1]{University of Jyväskylä, Faculty of Information Technology, Jyväskylä, Finland}
\affil[2]{Finnish Artificial Intelligence Research Network, Jyväskylä, Finland}
\affil[*]{corresponding.faprezja@jyu.fi}

\begin{abstract}
Machine learning (ML) research and application often involve time-consuming steps such as model architecture prototyping, feature selection, and dataset preparation. To support these tasks, we introduce the Deep Fast Machine Learning Utils (DFMLU) library, which provides tools designed to automate and enhance aspects of these processes. Compatible with frameworks like TensorFlow, Keras, and Scikit-learn, DFMLU offers functionalities that support model development and data handling. The library includes methods for dense neural network search, advanced feature selection, and utilities for data management and visualization of training outcomes. This manuscript presents an overview of DFMLU's functionalities, providing Python examples for each tool.

\end{abstract}

\begin{document}

\flushbottom
\maketitle
\thispagestyle{empty}

\section*{Introduction}

Machine Learning (ML) has steadily evolved, creating numerous tools and frameworks that support model development and deployment. Within the field of artificial intelligence, deep learning\cite{lecun2015deep} stands as a significant subset of ML that typically focuses on deep neural networks and big data. Deep architectures can enable learning complex patterns and representations from large data, allowing for developments in areas such as Biomedical computer vision\cite{kather2019predicting,bychkov2018deep,skrede2020deep,calimeri2017biomedical,prezja2022deepfake, paszke2019pytorch, ronneberger2015u,cciccek20163d, he2016deep,PREZJA2024e37561, bulten2019automated, duc20203d, prezja2023exploring}, and natural language processing\cite{kombrink2011recurrent,sutskever2014sequence,bahdanau2014neural,vaswani2023attentionneed,devlin2018bert,lan2019albert,hoffmann2022training,du2021glm,black2021gpt,he2020deberta,lewis2020retrieval,chowdhery2023palm,zaheer2020big}. Well-established libraries, including PyTorch\cite{paszke2019pytorch}, TensorFlow\cite{tensorflow2015-whitepaper}, Keras\cite{chollet2015keras}, and Scikit-learn\cite{scikit-learn_variance_threshold_code}, offer dependable platforms for building a variety of ML and deep learning models. As model architectures increase in complexity and datasets grow, tasks such as feature selection, architecture design, and data preparation may become more resource-intensive. The Deep Fast Machine Learning Utils (DFMLU) library is a Python-based utility collection designed to complement existing ML libraries by offering additional tools that simplify aspects of these tasks. DFMLU emphasizes efficiency in the ML workflow by partly automating processes such as dense neural network search, advanced feature selection, dataset management, and performance tracking.

\section*{Methods}
This section outlines the DFMLU library's key tools, focusing on model architecture search and advanced feature selection.

\subsubsection*{Principal Component Cascade Dense Neural Architecture Search (PCCDNAS)}

PCCDNAS provides an automated method for designing dense neural networks. This approach uses PCA (Principal Component Analysis), which systematically sets the number of neurons in each network layer. After applying PCA to the initial data, the neuron count for the first layer is determined based on the principal component counts (PCs) for a given variance threshold. Subsequently, the cascade mechanism ensures that the activations from a trained layer undergo PCA again. This process, in turn, determines the neuron count for the subsequent layers using the same principal component variance threshold criteria. PCCDNAS may reduce the need for manual dense architecture tuning in this context. While a more comprehensive description of the PCCDNAS method will be presented in future research, it can be viewed as a specialized case of additive auto-encoding\cite{karkkainen2023additive}, while both works were developed independently.

\textbf{PCCDNAS Pseudo-Algorithm}:
\begin{lstlisting}[language=Python]
# Pseudo-Algorithm for PCCDNAS:

1. Initialize:
   - Create an empty neural network model.
   - Create an empty list to store the number of neurons for each layer.

2. Data Initialization:
   - Accept training data and labels.
   - Center or normalize the data if required.

3. Initialize Model Search:
   - Set hyperparameters (e.g., number of layers, PCA variance threshold, etc.).

4. Build the Neural Network Model:
   - While not reached the desired number of layers:
     a. If at the first layer build stage, use the original training data.
     b. For subsequent layer build stages:
        - Train the model.
        - Extract the activations from the last layer for each data point.
     c. Perform PCA on the data (original or activations).
     d. Determine the number of principal components that meet the variance threshold.
     e. Set the number of neurons in the next layer based on the determined principal components count.
     f. Add the layer to the model.
\end{lstlisting}

\textbf{Implementation Example}:
\begin{lstlisting}[language=Python]
from deepfastmlu.model_search import PCCDNAS
from sklearn.model_selection import train_test_split

# Split the data into training and validation sets
X_train, X_val, y_train, y_val = train_test_split(X, y, test_size=0.2, random_state=42)

# Initialize the PCCDNAS object
pccdnas = PCCDNAS()

# Initialize data for the model search
pccdnas.data_init(X_train=X_train,
                  y_train=y_train,
                  validation=(X_val, y_val),
                  normalize=True,
                  unit=True)

# Initialize model search hyperparameters
pccdnas.initialize_model_search(
    epochs=10,  # Number of training epochs
    layers=3,  # Number of layers in the neural network
    activation='relu',  # Activation function for the layers
    pca_variance=[0.95,0.84,0.63],  # Desired explained variance for PCA for each layer
    loss='binary_crossentropy',  # Loss function
    optimizer='adam',  # Optimizer
    metrics=['accuracy'],  # List of metrics to be evaluated during training
    output_neurons=1,  # Number of neurons in the output layer
    out_activation='sigmoid',  # Activation function for the output layer
    stop_criteria='val_loss',  # Criteria for early stopping
    es_mode='min',  # Mode for early stopping (maximize the stop_criteria)
    dropout=0.2,  # Dropout rate for dropout layers
    regularize=('l2', 0.01),  # Regularization type ('l2') and value (0.01)
    batch_size=32,  # Batch size for training
    kernel_initializer='he_normal',  # Kernel initializer for the dense layers
    batch_norm=True,  # Whether to include batch normalization layers
    es_patience=5,  # Number of epochs with no improvement for early stopping
    verbose=1,  # Verbosity mode (1 = progress bar)
    learn_rate=0.001  # Learning rate for the optimizer
)
# Build the model
model, num_neurons = pccdnas.build()
print("Number of neurons in each layer:", num_neurons)
\end{lstlisting}

\subsubsection*{Adaptive Variance Threshold (AVT)}

Variance thresholding is a common feature selection method that may enhance machine learning model performance by removing features with low variance~\cite{scikit-learn_variance_threshold_code,kuhn2013applied}. It computes the variance of each feature and eliminates those with variance below a specified threshold. Adaptive Variance thresholding (AVT) is a feature selector that dynamically determines a variance threshold based on the provided percentile of the feature variances. Features with a variance below this threshold are dropped. AVT was first introduced and detailed in Prezja et al. \cite{prezja2023adaptive}.

\textbf{Implementation Example}:
\begin{lstlisting}[language=Python]
from deepfastmlu.feature_select import AdaptiveVarianceThreshold
from sklearn.model_selection import train_test_split

# Split the data into training and validation sets
X_train, X_val, y_train, y_val = train_test_split(
    X, y, test_size=0.2, random_state=42
)

# Initialize AdaptiveVarianceThreshold with desired percentile
avt = AdaptiveVarianceThreshold(percentile=1.5, verbose=True)

# Fit AVT to the training data
avt.fit(X_train)

# Transform both training and validation data
X_train_new = avt.transform(X_train)
X_val_new = avt.transform(X_val)
\end{lstlisting}

\subsubsection*{Rank Aggregated Feature Selection (RAFS)}
RankAggregatedFS is a feature selector that aggregates the rankings of features from multiple feature selection methods. It combines the scores or rankings of features from different methods to provide a unified ranking of features.

\textbf{Implementation Example}:
\begin{lstlisting}[language=Python]
from sklearn.model_selection import train_test_split
from sklearn.feature_selection import SelectKBest, mutual_info_classif, f_classif
from deepfastmlu.feature_select import RankAggregatedFS

# Split the data into training and test sets
X_train, X_test, y_train, y_test = train_test_split(
    X, y, test_size=0.2, random_state=42
)

# Define two feature selection methods
mi_selector = SelectKBest(score_func=mutual_info_classif, k=10)
f_classif_selector = SelectKBest(score_func=f_classif, k=10)

# Use RankAggregatedFS to aggregate rankings
rank_aggregated_fs = RankAggregatedFS(
    methods=[mi_selector, f_classif_selector], k=10
)

# Fit and transform the data using RAFS
X_train_new = rank_aggregated_fs.fit_transform(X_train, y_train)
X_test_new = rank_aggregated_fs.transform(X_test)
\end{lstlisting}

\subsubsection*{Chained Feature Selection (ChainedFS)}

ChainedFS is a feature selector that sequentially applies a list of feature selection methods. This approach enables the chaining of multiple feature selection techniques, where the output of one method becomes the input for the next. ChainedFS is useful when combining the strengths of different feature selection strategies or when a sequence of feature refinement operations is necessary.

\textbf{Implementation Example}:
\begin{lstlisting}[language=Python]
from sklearn.model_selection import train_test_split
from sklearn.pipeline import Pipeline
from deepfastmlu.feature_select import ChainedFS
from sklearn.feature_selection import VarianceThreshold, SelectKBest, mutual_info_classif

# Split the data into training and test sets
X_train, X_test, y_train, y_test = train_test_split(X, y, test_size=0.2, random_state=42)

# Create feature selection methods
variance_selector = VarianceThreshold(threshold=0.0)
k_best_selector = SelectKBest(score_func=mutual_info_classif, k=10)

# Initialize ChainedFS and create a pipeline
chained_fs = ChainedFS([variance_selector, k_best_selector])
pipeline = Pipeline([('feature_selection', chained_fs)])

# Fit the pipeline on the training data
pipeline.fit(X_train, y_train)

# Transform the training and test data using the pipeline
X_train_new = pipeline.transform(X_train)
X_test_new = pipeline.transform(X_test)
\end{lstlisting}

\subsubsection*{Mixing Feature Selection Approaches}

In some cases, combining multiple feature selection approaches might be desired. This section demonstrates how different feature selection techniques can be integrated into a single pipeline. In this example, we mix previously introduced approaches: Adaptive Variance Threshold (AVT) and Rank Aggregated Feature Selection (RAFS).

\textbf{Implementation Example}:
\begin{lstlisting}[language=Python]
from sklearn.model_selection import train_test_split
from sklearn.pipeline import Pipeline
from sklearn.feature_selection import SelectKBest, mutual_info_classif, f_classif
from sklearn.preprocessing import StandardScaler
from deepfastmlu.feature_select import RankAggregatedFS, AdaptiveVarianceThreshold

# Split the data into training and test sets
X_train, X_test, y_train, y_test = train_test_split(X, y, test_size=0.2, random_state=42)

# Create feature selection methods
adaptive_variance_selector = AdaptiveVarianceThreshold(percentile=1.5)
mi_selector = SelectKBest(score_func=mutual_info_classif, k=10)
f_classif_selector = SelectKBest(score_func=f_classif, k=10)

# Initialize RankAggregatedFS with multiple methods
rank_aggregated_fs = RankAggregatedFS(methods=[mi_selector, f_classif_selector], k=10)
pipeline = Pipeline([
    ('scaler', StandardScaler()),  # Normalize the data
    ('adaptive_variance_threshold', adaptive_variance_selector),  # Apply AdaptiveVarianceThreshold
    ('rank_aggregated_fs', rank_aggregated_fs)  # Apply RankAggregatedFS
])

# Fit the pipeline on the training data
pipeline.fit(X_train, y_train)

# Transform the training and test data using the pipeline
X_train_new = pipeline.transform(X_train)
X_test_new = pipeline.transform(X_test)
\end{lstlisting}

\subsubsection*{Dataset Splitter}
DFMLU also offers data management utilities, such as the Dataset Splitter, which streamlines the process of dividing datasets into training, validation, and test sets with stratification.

\textbf{Implementation Example}:
\begin{lstlisting}[language=Python]
from deepfastmlu.extra.data_helpers import DatasetSplitter

# Initialize the DatasetSplitter
splitter = DatasetSplitter(
    data_dir='path/to/dataset',
    destination_dir='path/to/splits',
    train_ratio=0.7,
    val_ratio=0.15,
    test_ratio=0.15,
    seed=42
)

# Run the splitting process
splitter.run()
\end{lstlisting}

\subsubsection*{Data Sub Sampler (miniaturize)}

The Data Sub Sampler is a utility class designed to sub-sample any repository-based dataset by selecting a fraction of files from the original dataset. This tool may be particularly useful when working with large datasets, allowing for the creation of a smaller dataset for quicker experimentation or debugging. The Data Sub Sampler provides a streamlined approach to 'miniaturizing' a dataset while preserving the structure of the original.

\textbf{Implementation Example}:
\begin{lstlisting}[language=Python]
from deepfastmlu.extra.data_helpers import DataSubSampler

# Define the paths to the original dataset and the destination directory for the subsampled dataset
data_dir = 'path/to/original/dataset'
subsampled_destination_dir = 'path/to/subsampled/dataset'

# Instantiate the DataSubSampler class with the desired fraction of files to sample
subsampler = DataSubSampler(data_dir, subsampled_destination_dir, fraction=0.5, seed=42)

# Create a smaller dataset by randomly sampling a fraction (in this case, 50%) of files from the original dataset
subsampler.create_miniature_dataset()
\end{lstlisting}

\subsubsection*{Plot Validation Curves}
The \texttt{plot\_history\_curves} tool helps visualize training and validation metrics across epochs. Showing the evolution of neural network performance aids in diagnosing underfitting, overfitting, and convergence patterns. This function can also highlight key values such as minimum loss, maximum accuracy or other metrics, to provide clearer performance insights.

\textbf{Implementation Example}:
\begin{lstlisting}[language=Python]
from deepfastmlu.extra.plot_helpers import plot_history_curves

# Training the model
history = model.fit(X_train, y_train_onehot, validation_data=(X_val, y_val_onehot), epochs=25, batch_size=32)

# Visualize the training history
plot_history_curves(history, show_min_max_plot=True, user_metric='accuracy')
\end{lstlisting}

\textbf{Example Output}:
\begin{figure}[ht]
    \centering
    \includegraphics[width=0.55\textwidth]{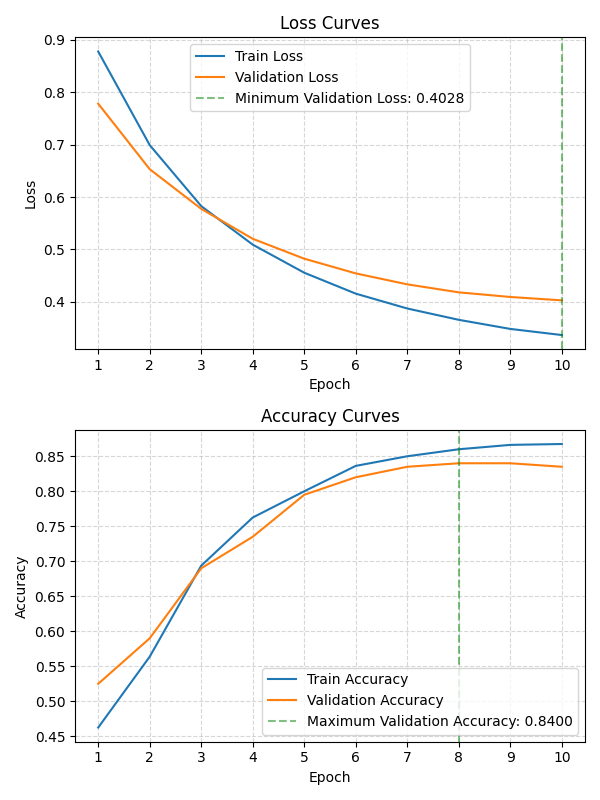}
    \caption{Validation Curves Plot}
    \label{fig:validation_curves}
\end{figure}

This visualization provides a clear look at how the accuracy and loss evolve over time.

\subsubsection*{Plot Generator Confusion Matrix}
The \texttt{plot\_confusion\_matrix} function is designed for generating confusion matrices from Keras image generators. This function automatically identifies class labels from the generator and provides a visual assessment of classification performance.

\textbf{Implementation Example}:

Below is an example of using \texttt{plot\_confusion\_matrix} in Python.

\begin{lstlisting}[language=Python]
from deepfastmlu.extra.plot_helpers import plot_confusion_matrix

# Create the confusion matrix for validation data
# model: A trained Keras model.
# val_generator: A Keras ImageDataGenerator used for validation.
# "Validation Data": Name of the generator, used in the plot title.
# "binary": Type of target labels ('binary' or 'categorical').
plot_confusion_matrix(model, val_generator, "Validation Data", "binary")
\end{lstlisting}

Data management and visualization tools derive from our Deep Fast Vision library\cite{prezja2023deep}.

\bibliography{main}

\section*{Data Availability}
The library and all related materials are available on the \href{https://github.com/fabprezja/deep-fast-machine-learning-utils}{official GitHub repository}.

\section*{Author contributions statement}
F. P. conceptualized and developed the library, and wrote the manuscript.

\end{document}